\documentclass[francais]{gretsi}
\usepackage[utf8]{inputenc}
\usepackage[T1]{fontenc}
\usepackage{amsmath, amssymb, amsfonts}
\usepackage[sorting=none,url=false,doi=false,eprint=false,isbn=false]{biblatex}
\usepackage{caption}

\begin{document}
\titre{Maximiser la marge pour une détection robuste des photomontages}

\auteurs{
  \auteur{Julien}{SIMON de KERGUNIC}{julien.simon@centrale.centralelille.fr}{1}
  \auteur{Rony}{Abecidan}{ronyabecidan@pm.me}{1,2}
  \auteur{Patrick}{Bas}{patrick.bas@cnrs.fr}{1}
  \auteur{Vincent}{Itier}{vincent.itier@imt-nord-europe.fr}{1,3}
}

\affils{
\affil{1}{Centre de Recherche en Informatique, Signal et Automatique de Lille, Avenue Henri Poincaré, 59655 Villeneuve d'Ascq, France } 
\affil{2}{Label4.ai, 111 avenue Victor Hugo 75016 Paris}
\affil{3}{IMT Nord Europe, Institut Mines-Télécom, Centre for Digital Systems, F-59000 Lille, France}
}

\resume{Malgré les progrès récents en détection de photomontages, les outils forensiques qui se basent sur l’apprentissage profond restent difficiles à exploiter en pratique, en raison de leur forte sensibilité aux données d'entraînement. Un simple post-traitement appliqué aux images d’évaluation peut suffire à dégrader leurs performances, compromettant leur fiabilité en contexte opérationnel. 
Dans cette étude, nous montrons qu’un même détecteur peut réagir différemment à des post-traitements inconnus selon les poids appris, malgré des performances similaires sur des données issues de la distribution d'entraînement. Ce phénomène s’explique par la variabilité des espaces latents induite par les entraînements, qui structurent différemment la séparation des classes.
Nos expériences révèlent une corrélation marquée entre la distribution des marges latentes et la capacité de généralisation du détecteur. Nous proposons ainsi une méthode simple pour une détection de photomontages robuste à destination des praticiens : entraîner plusieurs variantes d’un même modèle et sélectionner celle maximisant la marge dans l’espace latent, afin d’accroître la robustesse face aux post-traitements.
}

\abstract{Despite recent progress in splicing detection, deep learning-based forensic tools remain difficult to deploy in practice due to their high sensitivity to training conditions. Even mild post-processing applied to evaluation images can significantly degrade detector performance, raising concerns about their reliability in operational contexts.
In this work, we show that the same deep architecture can react very differently to unseen post-processing depending on the learned weights, despite achieving similar accuracy on in-distribution test data. This variability stems from differences in the latent spaces induced by training, which affect how samples are separated internally.
Our experiments reveal a strong correlation between the distribution of latent margins and a detector’s ability to generalize to post-processed images. Based on this observation, we propose a practical strategy for building more robust detectors: train several variants of the same model under different conditions, and select the one that maximizes latent margins.
}

\maketitle
\section{Introduction}

Un photomontage désigne toute modification structurée d’une image destinée
à en altérer le sens, qu’il s’agisse de l’insertion d’éléments externes
(\textit{splicing}) ou de la duplication interne de régions
(\textit{copy-move} ou \textit{cloning}).
Dans la suite, nous nous concentrons sur le scénario \textit{splicing}, cible privilégiée des détecteurs basés sur l’analyse du \textit{bruit résiduel}. Les détecteurs actuels, majoritairement fondés sur des réseaux de neurones profonds, recherchent des incohérences \textit{dans ce résidu}~\cite{bayar, trufor}. Néanmoins, leurs performances en contexte réel se révèlent souvent inférieures à celles rapportées dans la littérature. Cet écart s’explique par l’application de post-traitements inconnus aux images falsifiées : une chaîne de développement suffit à induire un \textit{décalage de domaine}, en modifiant la distribution statistique des images, authentiques comme falsifiées, et en dégradant la robustesse des détecteurs. Ces transformations (netteté, débruitage, etc.) altèrent le bruit et introduisent des dépendances entre zones originales et manipulées, rendant les photomontages moins détectables par les méthodes forensiques basées sur la détection d'anomalies locales. Par conséquent, tous les détecteurs de photomontages s'appuyant sur les résidus de bruits sont affectés par les post-traitements.

\paragraph{Robustesse opérationnelle des détecteurs.}

Dans la littérature forensique, le \textit{décalage de domaine} induit par les post-traitements est un problème connu. Les approches \textit{centrées sur les données} consistent à réfléchir à la construction d'une base d'entrainement pertinente pour maximiser la performance pratique des détecteurs ~\cite{domainrandom2,clevermixture,wifs2022,wifs2023}, tandis que les approches \textit{centrées sur le détecteur} visent à construire des détecteurs naturellement plus robustes face aux données hors distribution en utilisant par exemples des méthodes d'adaptation de domaines ~\cite{forensictransfer,advsforensics,wifs2021}. Ces méthodes reposent souvent sur des hypothèses fortes non validées en pratique comme l'équilibrage des classes en cible. À notre connaissance, la robustesse hors distribution sans accès aux cibles reste peu étudiée en détection de photomontage.

\begin{table}[htbp]
  \centering
  \resizebox{\columnwidth}{!}{%
    \begin{tabular}{|c|c|c|c|}
      \hline
      Graine & Précision sur la source & Précision moyenne sur les cibles & Écart-type sur les cibles \\ \hline
      4 & 84\% & 72\% & 1.8 \\ \hline
      6 & 84\% & 74\% & 2.0 \\ \hline
      8 & 84\% & 66\% & 2.3 \\ \hline
    \end{tabular}%
  }
  \caption{\scriptsize Impact de différentes graines d'initialisation sur les performances (hors distribution) du détecteur de Bayar~\cite{bayar}. La précision moyenne est obtenue en moyennant les performances d'un détecteur de Bayar, entraîné avec trois graines sur 20 cibles post-traitées via RawTherapee. Ensemble d'entraînement : $N_{\text{source}} \sim 20{,}000$ patches ; ensembles de test : $N_{\text{source}} \sim N_{\text{target}} \sim 7{,}000$ patches.}
  \label{tab:teasing}
\end{table}
\vspace{-0.2cm}
\noindent
Le tableau~\ref{tab:teasing} présente les performances d’un même détecteur de photomontage entraîné avec trois initialisations différentes. Bien que les performances soient similaires sur un ensemble d'images test suivant la même distribution que les images d'entraînement, les résultats varient grandement sur 20 versions post-traitées de ce même ensemble d'images. Cela montre que des entraînements différents qui conduisent à des performances équivalentes sur un ensemble de test suivant la distribution d'entraînement, peuvent mener à des modèles inégalement robustes sur des échantillons hors distribution, du fait de la convergence vers des minimums locaux distincts. Cette disparité soulève des questions sur les pratiques d’entraînement à adopter pour renforcer la robustesse des détecteurs de photomontages face aux post-traitements.

\paragraph{Contributions.}
Nous cherchons à comprendre pourquoi une même architecture peut se montrer plus ou moins
robuste aux post-traitements selon l'entraînement suivi. Dans un scénario réaliste,
un expert peut entraîner plusieurs détecteurs et choisir le plus robuste, sans connaître
la distribution d’évaluation. Pour cela, nous testons plusieurs entraînements d’un même modèle
et analysons leurs comportements hors distribution.

\vspace{0.2em} 
\begin{enumerate}
  \item Mise en évidence de l’effet négatif d’un surapprentissage à la source
        sur la généralisation du détecteur.
  \item Découverte d’une corrélation claire entre la répartition des
        échantillons d’entraînement dans les espaces latents et la robustesse
        du détecteur face aux post-traitements.
\end{enumerate}

La section~\ref{sec:Formalisation} précise notre cadre d’étude ; la
section~\ref{sec:margin} présente les expériences, et la
section~\ref{sec:conclusion} discute des résultats et perspectives.

\section{Formalisation}
\label{sec:Formalisation}
\subsection{Formulation du problème et scénario}

En adoptant les conventions de ~\cite{sepak}, une chaîne de traitement peut se représenter par un vecteur $\omega \in \Omega$ regroupant les paramètres d'une chaîne de développement (le coefficient de débruitage, le facteur de qualité JPEG, \textit{etc.}). Pour la détection de photomontages, il est courant d'utiliser des modèles issus de l'apprentissage automatique:
\begin{align*}
    f(x \mid \theta_{\omega}) : \ & \mathcal{X} \rightarrow \{\text{authentique}, \text{falsifié}\} \\
    & x \mapsto y.
\end{align*}
Ici, $\theta_{\omega} \in \Theta$ représente les paramètres appris à partir d’images authentiques et falsifiées ayant subi un post-traitement selon les paramètres $\omega$.  
Pour évaluer l’impact d’un décalage de domaine, il est courant de calculer l'\textit{écart de généralisation} entre une source $s$ (base d’entraînement) et une cible $t$ (base d’évaluation) :
{\normalsize
\begin{align*}
    \mathcal G_{f(x \mid \theta_{\omega})}(\omega_s,\omega_t) = 
    \ &\mathbb E_{(x,y) \sim P((x,y)| \omega_s)} \left( f(x \mid \theta_{\omega_s} ) = y \right) \\ 
    - \ &\mathbb E_{(x,y) \sim P((x,y)| \omega_t)} \left( f(x \mid \theta_{\omega_s} ) = y \right). \ \ (1)
\end{align*}
}
Cet écart correspond à la différence de performance entre un scénario idéal — où source et cible partagent le même post-traitement — et un scénario réaliste où la distribution cible est inconnue.  Dans notre cadre, les échantillons cibles ne sont pas accessibles à l’apprentissage ; l’objectif est donc de construire un détecteur de photomontages aussi robuste que possible à des post-traitements inconnus. 

\subsection{Marges dans les espaces latents}

En détection de photomontage, nous considérons deux classes : \textit{authentique} et \textit{falsifiée}.  
En conséquence, nos modèles produisent deux scores de logit, \( f_1 \) et \( f_2 \), pour chaque entrée \(x \in \mathcal{X}\).  
La classe prédite est celle ayant le score le plus élevé, c’est-à-dire \( i^* = \arg \max_i f_i(x) \).  
Les détecteurs sont constitués de couches successives, chaque couche projetant son entrée dans un nouvel espace latent.  
La frontière de décision linéaire dans l’espace latent final apparaît non linéaire dans les espaces latents précédents, menant à une frontière de décision spécifique à chaque espace latent.  
La frontière de décision $\mathcal{D}^l$ du \( l \)-ième espace latent de notre détecteur est définie comme l’ensemble des points $x^l$ de cet espace pour lequel le détecteur est incertain entre les deux classes :
$$
\mathcal{D}^l = \left\{ x^l \mid f_1(x^l) = f_2(x^l). \right\} \ \ (2)
$$
\noindent
On peut alors définir la marge d’un échantillon latent \(x^l\) par rapport à cette frontière $\mathcal{D}^l$ comme étant la plus petite perturbation \(\delta^l\) nécessaire pour amener \(x^l\) sur la frontière de décision de l’espace latent \(l\) :
\[
d^p_{f, x^l} = \min_{\boldsymbol{\delta}} \|\delta^l\|_p \quad \text{t.q.} \quad f_1(x^l + \delta^l) = f_2(x^l + \delta^l). \ \ (3)
\]

\noindent
L’écart de performance causé par les post-traitements résulte des biais spécifiques appris par les détecteurs de photomontages, dont les frontières de décision s’adaptent à la distribution source mais peinent à généraliser sur d'autres distributions. Intuitivement, des frontières trop proches des échantillons rendent le modèle plus facilement sensible aux variations induites par les post-traitements. Une étude précédente a justement montré une corrélation entre l’écart de généralisation et la distribution des marges latentes~\cite{margin}. Toutefois, cette analyse se limitait à deux cibles et à des classifieurs d'images au sens sémantique.

\section{Robustesse et marges latentes}
\label{sec:margin}
\subsection{Protocole expérimental}

\begin{figure}[t!]
  \centerline{
  \includegraphics[width=0.9\columnwidth]{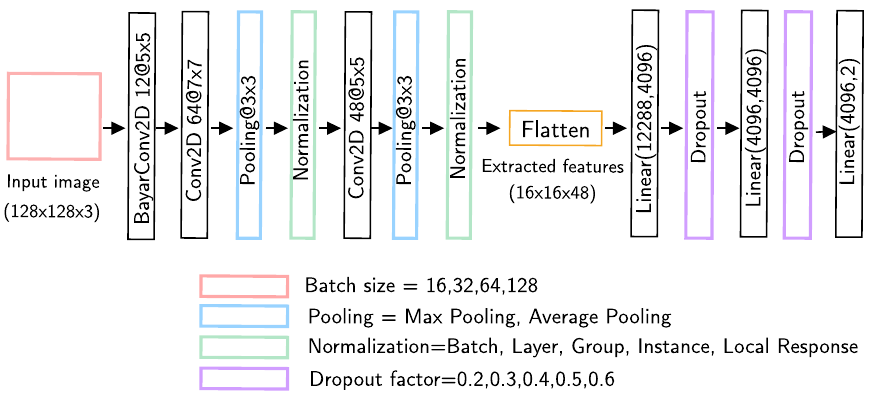}}
  \vspace*{-0.2cm}
  \caption{\scriptsize Schéma des détecteurs de Bayar~\cite{bayar}. Les cellules colorées indiquent les hyperparamètres ou opérateurs modifiés au cours de nos 200 entraînements.}
  \label{fig:bayar}
\end{figure}

\paragraph{Choix du détecteur et hyperparamètres.}
Nos expériences s’appuient sur le détecteur de photomontages de Bayar et Stamm~\cite{bayar}, un réseau de neurones convolutionnel largement utilisé par la communauté forensique. Son architecture, illustrée dans la figure~\ref{fig:bayar}, suit un schéma classique (Convolution + Max Pooling + Couches entièrement connectées), à l’exception de la première couche convolutionnelle, contrainte à effectuer un filtrage passe-haut :
\begin{center}
	\vspace{-0.5cm}
	\footnotesize{
		$$
\left\{\begin{array}{l}
\boldsymbol{w}_{k}^{(1)}(0,0)=-1, \\
\sum_{m, n \neq 0} \boldsymbol{w}_{k}^{(1)}(m, n)=1.
\end{array}\right.
$$
	}
\end{center}
Cette contrainte favorise l’extraction de résidus de bruit trahissant des manipulations. Dans cet article, nous utilisons cette architecture pour analyser son comportement face à 20 cibles ayant subi des post-traitements variés. Les choix suivants ont été adoptés pour l’optimisation et les hyperparamètres :

\begin{itemize}
\setlength{\itemsep}{-0.1em}
    \item L’entraînement s’effectue sur un maximum de 115 époques, suffisant pour assurer la convergence.
	\item L’optimiseur utilisé est SGD.
    \item Le \textit{batch size} est fixé à 128, un compromis adapté aux capacités d'un GPU classique et à la stabilité de l’apprentissage.
    \item Le \textit{learning rate} initial est de $10^{-3}$, réduit d’un facteur 10 après 4 époques sans amélioration.
    \item Les poids sont initialisés selon les réglages par défaut de \texttt{PyTorch}~\cite{he2015delving}.
	\item Toutes les expériences utilisent la graine 22 (choisie aléatoirement) pour assurer la reproductibilité et comparabilité des expériences.
\end{itemize}
Les ensembles source et cible sont divisés en sous-ensembles entraînement/validation/test selon la proportion $0.6/0.2/0.2$. Un mécanisme d'\textit{early stopping}, basé sur la précision de l'ensemble de validation est utilisé pour éviter le sur-apprentissage.

\paragraph{Construction des domaines source et cible.}
Nous utilisons DEFACTO~\cite{DEFACTODataset}, le plus grand jeu de données public dédié à la détection de photomontages, en raison  du réalisme des images et du contrôle offert sur les post-traitements via le format TIFF. La catégorie photomontage est divisée en deux sous-ensembles indépendants de taille équivalente : l’un pour la source, l’autre pour les cibles.
Chaque image est découpée en patches de $128 \times 128$ pour un entraînement homogène. Un patch est étiqueté \textit{falsifié} si la zone falsifiée couvre entre 10\,\% et 40\,\% de sa surface, une plage choisie pour éviter les cas extrêmes où le détecteur aurait du mal à distinguer les distributions de bruit. Des patchs \textit{authentiques} sont ensuite ajoutés pour équilibrer les classes. Ce prétraitement génère environ 20\,000 patches pour l'entraînement et 7\,000 pour le test. La source est constituée uniquement d’images TIF originales, tandis que les cibles subissent divers post-traitements, appliqués directement sur les TIF avant découpage afin d’éviter les artefacts. Enfin, les cibles partagent un contenu visuel similaire, ce qui permet d’attribuer les écarts de performance uniquement aux effets des post-traitements, et non aux différences de contenu.

\paragraph{Pipelines de post-traitement.}
Nous définissons 20 chaînes de développement en combinant des opérations de débruitage par ondelettes et de renforcement de netteté via \href{https://www.rawtherapee.com/}{RawTherapee}. Une compression JPEG avec un facteur de qualité de 70 est ensuite appliquée en fin de pipeline à l’aide de \href{https://www.imagemagick.com/}{ImageMagick}, assurant un contrôle complet sur le processus. Ce choix constitue un bon compromis entre réalisme et variabilité des cibles. 



\paragraph{Entraînements multiples d’une même architecture.}
Nous proposons ici de faire varier les hyperparamètres et opérateurs du détecteur de Bayar afin d'étudier leur influence sur la capacité du modèle à généraliser à des échantillons post-traités.  
Nous réalisons 200 entraînements du détecteur de Bayar en modifiant la taille de batch, le type de pooling, la normalisation et le dropout, selon le schéma présenté en figure~\ref{fig:bayar}.

\paragraph{Analyse par courbes de quantiles.}
Les courbes de quantiles permettent de visualiser comment la distribution de l'écart de généralisation évolue dans une fenêtre glissante centrée sur des valeurs successives de diverses métriques.  
Pour chaque courbe de quantile, nous explorons les valeurs de la métrique avec un pas donné et une taille de fenêtre spécifiée (indiquée dans les légendes), en équilibrant le compromis entre localisation précise et fiabilité de l’estimation des quantiles.

\subsection{Surapprentissage sur la source}

Les 200 entraînements confirment le phénomène de surapprentissage à la source, évoqué dans~\cite{margin} : plus l'accuracy sur l'ensemble de test source est élevée, plus l’écart de généralisation vers les cibles augmente. Ce comportement suggère que le réseau apprend progressivement des biais spécifiques à la source, optimisant ses performances au détriment de la robustesse sur les cibles. 



\subsection{Marges latentes et écarts de généralisation}
\begin{figure}[ht!]
  \centerline{
  \includegraphics[width=0.8\columnwidth]{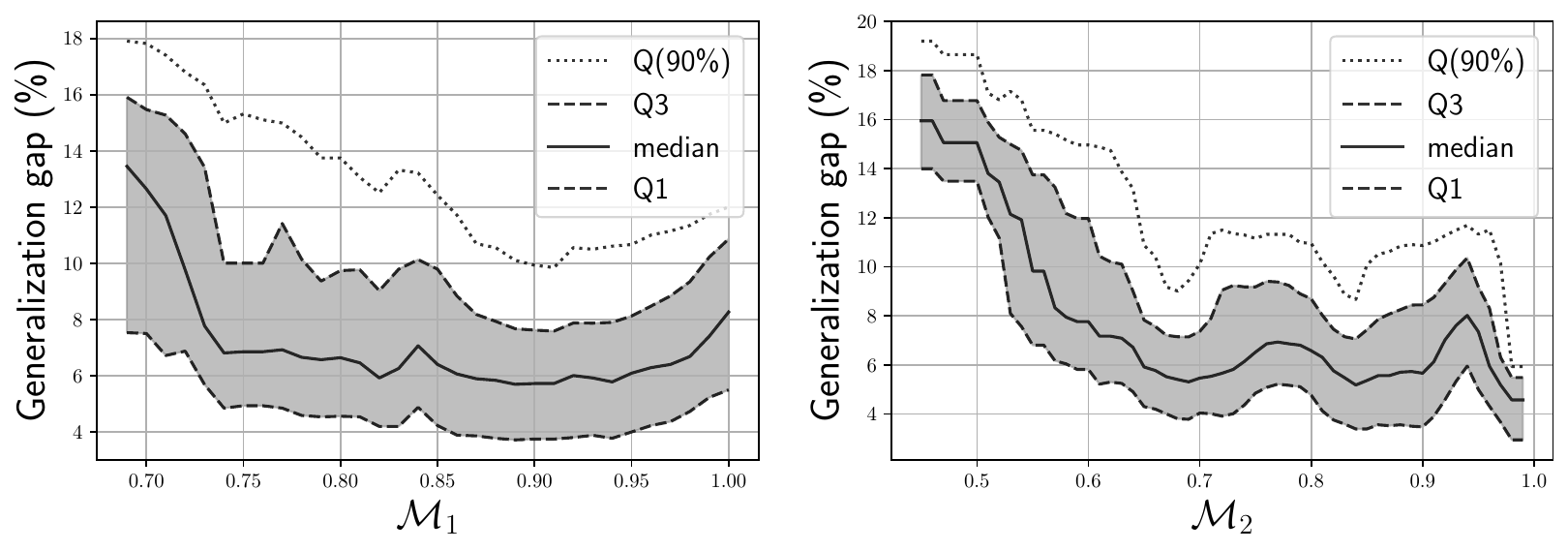}
    }
  \vspace*{-0.3cm}
  \caption{\scriptsize Courbes de quantiles représentant l’évolution de l'écart de généralisation sur nos 20 domaines en fonction des métriques de marge $\mathcal{M}_1$ et $\mathcal{M}_2$, calculées à partir des marges latentes issues de toutes les couches. Ces métriques sont normalisées pour comparaison. Q1 est le premier quartile, Q3 est le troisième quartile et Q(90\%) est le 90e percentile. Les points métriques sont explorés avec un pas de 0{,}01 et une taille de fenêtre de 0{,}1. \\ \hspace*{2.5cm}}
  \label{fig:margin_gp}
\end{figure}

Nous cherchons à vérifier s’il existe une corrélation entre les marges latentes des échantillons d'entrainement et les écarts de généralisation. Pour éviter les biais dus au sous-apprentissage, l’analyse est restreinte aux 138 modèles ayant atteint au moins 75\,\% de précision sur la source. Le calcul des marges latentes suit une méthodologie inspirée de~\cite{margin} :

\begin{enumerate}
\setlength{\itemsep}{-0.1em}
  \item Estimer les marges latentes $d^2_{f, x^l}$ de chaque échantillon source $x^l$ à l’aide des logits $f_1$, $f_2$ et de leurs gradients par rapport à chaque couche, en prenant soin de les normaliser pour assurer l’indépendance à l’échelle. Nous excluons les marges négatives causées par les erreurs de classification. Davantage de détails sur ce calcul sont disponibles dans \cite{margin}. 
  \item Résumer les distributions de marges pour chaque espace latent à l’aide de vecteurs $\boldsymbol{\mu_l}$ contenant des statistiques descriptives (premier et troisième quartiles, médiane, bornes supérieure et inférieure). Il est ensuite possible de concaténer tous les $\boldsymbol{\mu_l}$ dans un unique vecteur $\boldsymbol{\mu}$.
  \item Combiner les statistiques de marge de chaque vecteur afin de dériver une métrique de marge $\mathcal{M}$. Nous suggérons de tester des combinaisons simples en calculant la somme des statistiques élevées à une certaine puissance $\alpha$ : 
  \[
  \mathcal{M}_{\alpha} = \sum_i \mu_i^{\alpha}. \ \ (4)
  \]
  L’élévation à la puissance $\alpha$ permet de mieux accentuer les différences de marge entre architectures.
\end{enumerate}
\noindent
Contrairement à~\cite{margin}, notre objectif n’est pas de prédire l’écart de généralisation à partir des marges, mais de fournir aux analystes forensiques une métrique pour évaluer la qualité de l’entraînement en vue d’une détection robuste de photomontage. 
 Pour chaque détecteur de Bayar $f^i(x|\theta_{s})$ entraîné sur notre source $s$ et chaque cible $t$ ayant subi un post-traitement via le pipeline $\omega_t$, nous générons des couples :  
\[
[\mathcal{M}_{\alpha}[f^i(x|\theta_{s})] \ , \ \mathcal G_{f^i(x|\theta_{\omega_{s}})}(\omega_{s},\omega_t)]. \ \ (5)
\]
Les 2760 couples obtenus permettent d’étudier la corrélation entre $\mathcal{M}_{\alpha}$ et les écarts de généralisation. Les courbes de quantiles de la figure~\ref{fig:margin_gp} montrent que les détecteurs les plus robustes aux post-traitements sont ceux qui séparent le mieux leurs échantillons d’entraînement dans l’espace latent, en particulier avec $\mathcal{M}_{2}$.

\vspace{-0.8cm}
\subsection{Importance de chaque marge latente}

Bien que~\cite{margin} souligne que l’analyse d’un seul espace latent soit insuffisante pour expliquer l’écart de généralisation, nous explorons ici la corrélation entre $\mathcal{G}$ et $\mathcal{M}_1$ en calculant les marges couche par couche.  
Ce choix est motivé par l’observation d’une corrélation positive entre ces deux métriques dans les cas de marges élevées, que nous cherchons à mieux comprendre.

\begin{figure}[ht!]
  \centerline{
  \includegraphics[width=0.8\columnwidth]{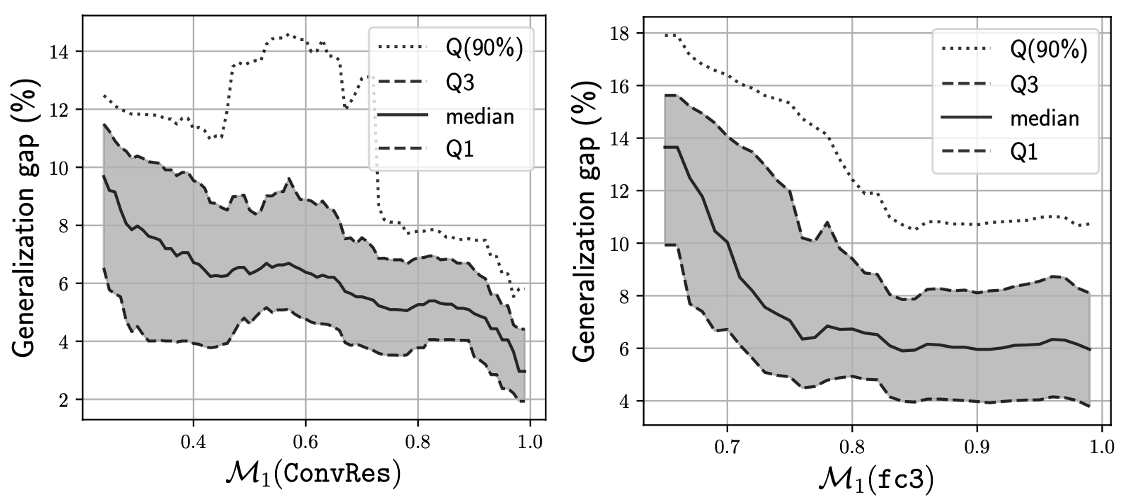}
    }
  \vspace*{-0.3cm}
  \caption{\scriptsize Courbes de quantiles représentant l’évolution de l'écart de généralisation sur nos 20 domaines en fonction de $\mathcal{M}_1$, calculée à partir des marges latentes associées à la première couche (ConvRes) et dernière (fc3) de nos détecteurs de Bayar. Ces métriques sont normalisées pour comparaison. Q1 est le premier quartile, Q3 est le troisième quartile et Q(90\%) est le 90e percentile. Les points métriques sont explorés avec un pas de 0{,}01 et une taille de fenêtre de 0{,}1.}
  \label{fig:first_last_layers_margin}
\end{figure}
\noindent
La figure~\ref{fig:first_last_layers_margin} montre une corrélation nette pour la première et la dernière couche latente. En revanche, nos expériences ne nous ont pas montré de corrélation dans les couches intermédiaires. Nous expliquons l’effet observé sur la première couche par le rôle des couches en amont, qui capturent des caractéristiques générales~\cite{upstream}: des marges élevées à ce niveau contribuent à la robustesse sur la cible. En sortie, la dernière couche étant la plus spécifique, de faibles marges y rendent le modèle sensible aux perturbations induites par les post-traitements. D’où l’intérêt d’un espace final bien séparateur, comme le suggèrent les approches basées sur les pertes contrastives~\cite{contrastive}. L’absence de corrélation dans les couches intermédiaires s’explique sans doute par leur rôle transitoire : elles visent à préparer la séparation finale, sans structurer leur propre espace latent.



\section{Conclusion}
\label{sec:conclusion}

Cet article met en lumière la sensibilité des détecteurs de photomontage aux post-traitements inconnus, en montrant que des entraînements distincts d’une même architecture peuvent conduire à des capacités de généralisation très variables. Nos résultats soulignent qu’un surapprentissage à la source dégrade la robustesse hors distribution, justifiant l’utilisation de critères d’arrêt précoces adaptés.
Nous proposons alors une métrique de marge latente, construite à partir de statistiques simples, et corrélée à l’écart de généralisation. Les marges issues des premières et dernière couches latentes sont particulièrement informatives. 
Nous recommandons ainsi d'entraîner plusieurs variantes d’un détecteur et de sélectionner celle maximisant les marges dans les couches clés.  
Dans nos travaux futurs, nous étendrons cette analyse à d’autres détecteurs et explorerons la conception d’architectures résilientes via des pertes contrastives, tout en étudiant le rôle des hyperparamètres dans la généralisation hors distribution.

\section{Remerciements}
Ces travaux ont bénéficié d’un accès aux moyens de calcul de l’IDRIS au travers de l'allocation de ressources 2025-AD011016555 attribuée par GENCI.
\label{sec:Remerciement}

\AtNextBibliography{\footnotesize} 
{
\printbibliography
}

\end{document}